\documentclass{article}

\PassOptionsToPackage{numbers, compress}{natbib}
\usepackage[preprint]{neurips_2026}
\makeatletter
\renewcommand{\@noticestring}{}
\makeatother

\usepackage[utf8]{inputenc}
\usepackage[T1]{fontenc}
\usepackage{hyperref}
\usepackage{url}
\usepackage{booktabs}
\usepackage{amsfonts}
\usepackage{nicefrac}
\usepackage{microtype}
\usepackage{xcolor}
\usepackage{amsmath}
\usepackage{amssymb}
\usepackage{graphicx}
\usepackage{subcaption}
\usepackage{multirow}
\usepackage{wrapfig}
\usepackage[capitalize,noabbrev]{cleveref}

\usepackage{enumitem}

\usepackage{tikz}
\usetikzlibrary{arrows.meta,positioning,fit,calc}

\definecolor{LinkBlue}{HTML}{1A4FA3}
\hypersetup{colorlinks, linkcolor={LinkBlue}, citecolor={LinkBlue}, urlcolor={LinkBlue}}
\usepackage[labelfont=bf]{caption}

\newcommand{\oursTD}{\ensuremath{\mathrm{SA}_{\mathrm{TD}}}}
\newcommand{\oursBU}{\ensuremath{\mathrm{SA}_{\mathrm{BU}}}}

\newcommand{\oursTDb}{\ensuremath{\mathbf{SA}_{\mathbf{TD}}}}
\newcommand{\oursBUb}{\ensuremath{\mathbf{SA}_{\mathbf{BU}}}}

\newcommand{\ours}{\textsc{SA}}

\title{Object-Centric Learning for Part-Whole Hierarchies:\\
       Promises, Limitations, and Open Challenges}

\author{%
  Amir Mohammad Karimi Mamaghan\textsuperscript{1}\thanks{Corresponding author: \href{mailto:amkm@kth.se}{\texttt{amkm@kth.se}}.}\quad
  Ferdinand Kapl\textsuperscript{2,3,4}\\
  \textbf{Andrea Dittadi\textsuperscript{2,3}}\quad
  \textbf{Karl Henrik Johansson\textsuperscript{1,5}}\quad
  \textbf{Stefan Bauer\textsuperscript{2,3,4}}\\[6pt]
  {\normalfont\small\textsuperscript{1}KTH Royal Institute of Technology}\\
  {\normalfont\small\textsuperscript{2}Helmholtz AI, Munich\quad
  \textsuperscript{3}Technical University of Munich}\\
  {\normalfont\small\textsuperscript{4}Munich Center for Machine Learning (MCML)\quad
  \textsuperscript{5}Digital Futures}
}

\hypersetup{%
  pdftitle={Object-Centric Learning for Part-Whole Hierarchies: Promises, Limitations, and Open Challenges},
  pdfauthor={Amir Mohammad Karimi Mamaghan, Ferdinand Kapl, Andrea Dittadi, Karl Henrik Johansson, Stefan Bauer},
  pdfsubject={Object-centric learning and part-whole representations},
  pdfkeywords={object-centric learning, part-whole hierarchies, Slot Attention}
}

\begin{document}

\maketitle

\begin{abstract}
Object-centric (OC) learning aims to decompose visual scenes into reusable entity-like representations, making it a natural candidate for modeling part-whole structure. However, most OC methods produce a flat set of slots rather than an explicit hierarchy. We study this gap through two hierarchical Slot Attention variants: a top-down model that first forms object-level slots and then extracts finer slots within each object region, and a bottom-up model that first forms part-level slots and then groups them into object-level slots. We evaluate the resulting representations on captioning and visual question answering (VQA) tasks and observe that hierarchical slots can provide useful multi-level representations, with modest captioning gains and clearer VQA gains for the top-down model. However, hierarchy does not by itself provide a general advantage in downstream tasks, and its benefits depend on the architecture, representation level, task, and backbone. These models also show limited alignment with annotated parts and their parent relationships despite their downstream usefulness. We use these results as a case study to discuss the broader limitations of current OC learning for part-whole modeling.

\end{abstract}

\section{Introduction}
\label{sec:intro}

Visual recognition is fundamentally compositional. A scene contains objects, each object contains parts, and parts combine through spatial and functional relations that persist across viewpoints, occlusions, and contexts. This view has a long history in both neuroscience and machine learning. In the early visual cortex, simple oriented responses are progressively combined into more complex visual structure \citep{riesenhuber1999hierarchical,dicarlo2007untangling}. The ventral ``what'' stream supports increasingly invariant representations of shape and identity, while the dorsal ``where/how'' stream processes spatial organization and action-relevant structure \citep{goodale1992separate}. At the same time, lateral and top-down interactions help group local evidence into coherent percepts rather than isolated image patches \citep{riesenhuber1996neural}. These biological and cognitive observations have motivated many computational models of hierarchy, including connectionist parse-tree representations \citep{hinton1990mapping}, HMAX-style hierarchical recognition models \citep{riesenhuber1999hierarchical}, capsule networks \citep{sabour2017capsules}, and more recent proposals such as GLOM \citep{hinton2023glom}.

Object-centric (OC) learning provides a natural modern inductive bias for this problem. Instead of representing an image solely as a dense grid of features, OC methods aim to bind visual content to a small set of reusable, entity-like slots \citep{locatello2020object,greff2020binding}. This makes them appealing for compositional generalization, interpretability, and reasoning \citep{mamaghan2025exploring,kapl2026object,seitzer2022bridging}. Recent methods, such as DINOSAUR \citep{seitzer2022bridging}, show that Slot Attention applied to frozen self-supervised ViT features can produce object-like segmentations on complex natural images. However, the dominant OC formulation remains flat: an image is represented by a single set of slots, roughly corresponding to objects or regions. Recent works have started to move beyond this formulation through part-whole consistency objectives, hierarchical slot architectures, and multi-granularity decompositions \citep{wang2023cyclicwalks,maus2025efficient,dai2026cgsa,madan2026hsa}. These works demonstrate that hierarchical structure can be introduced into object-centric representations, but leave open how such hierarchies should be constructed and whether their different levels form useful semantic representations. In particular, it remains unclear when explicit hierarchy improves downstream representations and how closely its learned groupings correspond to annotated parts and their relationships.

To analyze this, we implement two simple part-whole variants of Slot Attention. The first follows a top-down design: it first discovers object-level slots and then applies a second Slot Attention module within each object region to discover parts. The second follows a bottom-up design: it first discovers part-level slots and then groups them into object-level slots. We first train the OC encoders to reconstruct frozen visual features, then freeze them and fine-tune a language model and visual adapter for each task. We evaluate these hierarchical representations on COCO captioning \citep{lin2014microsoft,chen2015cococaptions} and GQA visual question answering \citep{hudson2019gqa} and qualitatively inspect the learned masks. We also assess localization and parent binding on PACO \citep{ramanathan2023paco} to examine their correspondence to annotated part-whole structure. Our main finding is that adding hierarchy to Slot Attention does not, by itself, provide a general downstream advantage. The top-down model yields modest captioning gains and clearer improvements on GQA, while the bottom-up model is less reliable. The choice of hierarchy level also matters, with parts-only inputs sometimes matching or exceeding all-level inputs. Together, the downstream results and limited localization coverage on PACO suggest that useful multi-level representations do not necessarily recover annotated part-whole structure. We use this distinction to motivate a broader discussion of the promises and limitations of OC learning.

Overall, our contributions are as follows:
\begin{itemize}[topsep=0pt, parsep=0pt, partopsep=0em, left=1em]
\item We present two part-whole Slot Attention (SA) variants, \oursTD{} and \oursBU{}, to compare top-down and bottom-up hierarchies under a common reconstruction-based training setup.
\item We evaluate these representations against the flat baseline on COCO captioning and GQA visual question answering, using object-segmentation metrics, captioning and question-answering performance, a part-whole localization and binding analysis on PACO, and qualitative mask inspection. After observing conditional downstream and part-whole benefits from hierarchy, we discuss the limitations and challenges that current OC methods face for modeling part-whole hierarchies.\looseness=-1
\end{itemize}

\section{Background}
\label{sec:related}

\textbf{Object-centric learning.}
Object-centric (OC) learning aims to represent visual scenes as compositions of entity-like variables, and has been explored through recurrent attention, generative scene decomposition, spatial mixture models, video-based object discovery, contrastive learning, and slot-based architectures \citep{gregor2015draw, eslami2016attend, greff2017neural, burgess2019monet, engelcke2019genesis, engelcke2021genesis, kosiorek2018sequential, crawford2019spatially, jiang2019scalor, lin2020space, locatello2020object, kipf2019contrastive, kipf2022conditional, goyal2019recurrent, wu2021generative, sajjadi2022object, elsayed2022savi++, singh2021structured, singh2022simple, singh2021illiterate, seitzer2022bridging, jiang2023object, wu2023slotdiffusion, jabri2024dorsal, daniel2023ddlp, daniel2022unsupervised}. Among these methods, Slot Attention \citep{locatello2020object} has become a widely used building block, with later work improving the grouping module \citep{biza2023invariant, jia2023improving, majellaro2024explicitly}, replacing reconstruction decoders with transformer-based decoders \citep{singh2022simple,singh2021illiterate}, combining slots with pretrained visual backbones \citep{seitzer2022bridging}, or integrating slots with diffusion models \citep{jiang2023object, wu2023slotdiffusion, jabri2024dorsal}. Beyond segmentation, recent work has also revisited how OC representations should be evaluated, including visual reasoning, reinforcement learning, OOD generalization, video understanding, and robotic planning \citep{Ding2021, wu2022slotformer, driess2023palm, Dittadi2021, papa2022inductive, weis2021benchmarking, yoon2023investigation, yang2024benchmarking, haramati2024entity, montero2024successes, rubinstein2025donewithoc, baldassarre2024clustering, mamaghan2025exploring, kapl2026object,singh2026evaluating}. Despite these advances, the dominant formulation remains single-level: each image is represented as a flat set of slots.

\textbf{Hierarchical object-centric learning.}
Some recent works move beyond a single flat set of slots, although they target different notions of hierarchy. \citet{wang2023cyclicwalks} model part-whole consistency between ViT patch features and object slots, while Interpreter \citep{maus2025efficient} introduces a hierarchical Slot Attention architecture for video. CGSA \citep{dai2026cgsa} uses coarse-to-fine hierarchical slots for source-free object detection, and \citet{madan2026hsa} learn holistic, semantic, and panoptic levels using partial supervision and hierarchical alignment. Other work studies whether hierarchy is already implicit in flat slot representations \citep{madan2026hyperbolic}. However, evaluation in most of these works is still largely based on object discovery or segmentation metrics, which prior studies show do not necessarily correlate with downstream representation quality \citep{mamaghan2025exploring,rubinstein2025donewithoc,singh2026evaluating}. Our focus is therefore complementary: we compare object-to-part and part-to-object hierarchies and evaluate them using unsupervised object-discovery metrics, downstream vision-language tasks, and an explicit part-whole evaluation.

\textbf{Part-whole hierarchy models.}
The idea of representing objects as compositions of parts predates modern OC models, with early neural and computational accounts studying how local features can be progressively combined into higher-level structures \citep{hinton1990mapping,riesenhuber1999hierarchical,dayan2006images}. Capsule networks make this explicit through routing mechanisms that bind parts to wholes \citep{sabour2017capsules}. More recently, GLOM proposed hierarchical columns with bottom-up, top-down, and lateral communication to form spatially coherent ``islands of agreement'' \citep{hinton2023glom}. Agglomerator and Agglomerator++ provide practical GLOM-inspired models for hierarchical patch grouping \citep{garau2022agglomerator,garau2024agglomeratorpp}, while HiLite uses similar bottom-up, top-down, and consensus updates to produce visualizable hierarchical groupings \citep{hilite2024}. Visual Parser and recursive image grammar models also study structured image parsing and compositional scene representations \citep{sun2021visualparser,fisher2023rnp}. Together, these works motivate hierarchical grouping, but robust and semantically meaningful part-whole modeling on complex real-world images remains challenging.

\section{Part-Whole Slot Attention}
\label{sec:method}

We develop part-whole Slot Attention to test whether OC representations can be extended from a flat set of slots to a simple two-level part-whole hierarchy. Given an image, a frozen backbone produces patch features $\mathbf{x}\in\mathbb{R}^{N\times C}$. Each hierarchy level applies a Slot Attention module to a set of input tokens and produces a fixed number of slots, along with attention masks over its inputs. Following common OC frameworks \citep{locatello2020object,seitzer2022bridging}, each level also has a lightweight patch decoder that maps its slots back to the original backbone feature space. We call this image-only feature-reconstruction stage \textit{upstream training}: it learns the slot modules and decoders while keeping the visual backbone frozen. Its objective is the sum of the mean-squared reconstruction losses from all hierarchy levels. In our experiments, we use two levels, but in principle, it can be extended to many more. We also insert a shallow transformer encoder block over the slots after each Slot Attention module at each level, allowing slots at the same level to communicate and exchange information through self-attention before decoding or before being passed to the next level \citep{hinton2023glom}. We develop two part-whole \ours{} variants, described below.

\textbf{Top-down \ours{}.}
The top-down variant, denoted \oursTD{}, follows an ``objects first, parts second'' design. The first Slot Attention module is applied to the full set of backbone patch features, producing coarse object-level slots and masks. Each object mask is then used to gate the original patch features, producing a masked feature set for each discovered object. A second Slot Attention module, shared across objects, is applied to these masked feature sets to produce part-level slots. Thus, part discovery is conditioned on the object-level masks, while still reconstructing the original backbone features at both levels. This makes the hierarchy easy to interpret, since each part slot is associated with an object-level parent, but it also makes the method more rigid and computationally expensive because the part-level module is applied separately for each object region.

\textbf{Bottom-up \ours{}.}
The bottom-up variant, denoted \oursBU{}, follows a ``parts first, objects second'' design. The first Slot Attention module is applied to the full set of backbone patch features, producing part-level slots. The second Slot Attention module then takes these part slots, rather than the original patch features, as input and groups them into a smaller set of object-level slots. The attention masks at the second level, therefore, describe assignments from part slots to object slots. As in the top-down model, each level is decoded with a separate decoder to reconstruct the original backbone features. This design is computationally simpler than \oursTD{}, but the object-level representation is only indirectly connected to the image through the quality and information content of the part-level slots.

After upstream training, the part-whole \ours{} encoder is frozen, and its slots are used as visual inputs to a language model fine-tuned on COCO captioning and GQA visual question answering. We call this supervised stage \textit{downstream training}. It fine-tunes the language model and visual adapter using caption or question-answer targets. For each hierarchy, we compare three inputs to the visual adapter: object slots alone, part slots alone, and both levels concatenated. \textbf{Object-level} uses only the slots intended to represent objects: the first-level slots for \oursTD{} and the second-level slots for \oursBU{}. \textbf{Part-level} uses only the slots intended to represent parts: the second-level slots for \oursTD{} and the first-level slots for \oursBU{}. \textbf{All-levels} concatenates slots from both hierarchy levels. This comparison tests whether adding hierarchy improves on a flat slot baseline and which hierarchy levels are useful for each task. For more implementation details, see \cref{app:arch,app:downstream}.\looseness=-1

\section{Experiments}
\label{sec:experiments}

\textbf{Setup.}
We train all OC encoders on the image sets used for COCO captioning and GQA separately, using frozen DINOv2-S \citep{oquab2023dinov2} features as the main visual input. As the flat OC baseline, we train DINOSAUR \citep{seitzer2022bridging} under the same setting, using the same frozen backbone features as the corresponding part-whole \ours{} models. We also run additional backbone experiments with DINOv3-S and DINOv3-L \citep{simeoni2025dinov3}, which are reported in \cref{app:dinov3}. Following common OC evaluation metrics, we evaluate unsupervised object discovery before language-model fine-tuning and report foreground-adjusted Rand index (FG-ARI) and mean best overlap (mBO) on COCO instance annotations.

\begin{wraptable}{R}{0.49\textwidth}
\centering
\captionsetup{font=small,justification=raggedright,singlelinecheck=false}
\caption{Unsupervised Object Discovery with DINOv2-S on COCO instance annotations. Parentheses indicate flat slot counts. Values are mean $\pm$ sample standard deviation over three upstream seeds. Bold indicates the best result in each column.}
\label{tab:main_discovery}
\smallskip
\fontsize{8.5}{10}\selectfont
\setlength{\tabcolsep}{1.5pt}
\renewcommand{\arraystretch}{1.10}
\begin{tabular*}{\linewidth}{@{\extracolsep{\fill}}lcc@{}}
\toprule
Model & FG-ARI $\uparrow$ & mBO $\uparrow$ \\
\midrule
DINOSAUR (7) & $39.92{\pm}0.24$ & $\mathbf{28.01{\pm}0.05}$ \\
DINOSAUR (35) & $20.31{\pm}0.16$ & $22.94{\pm}0.03$ \\
\oursTD{} & $37.61{\pm}0.42$ & $26.29{\pm}0.02$ \\
\oursBU{} & $\mathbf{41.30{\pm}0.87}$ & $27.86{\pm}0.22$ \\
\bottomrule
\end{tabular*}
\end{wraptable}

For downstream evaluation, we follow the BLIP-3 setup \citep{xue2024xgen} with Phi-3-mini-4k \citep{abdin2024phi3} as the language model, replacing the original vision backbone with our trained OC encoders. We freeze the complete OC encoder and jointly fine-tune the language model and a Perceiver \citep{jaegle2021perceiver} adapter with 64 output visual tokens. The adapter maps the selected slots into language-model input tokens. The model then predicts a caption or answer conditioned on these tokens and the text prompt. Unless specified otherwise, we report mean and sample standard deviation over three upstream seeds. We follow the COCO caption evaluation protocol \citep{chen2015cococaptions}, applying Stanford CoreNLP's Penn Treebank tokenizer \citep{manning2014corenlp} to predictions and references before computing CIDEr \citep{vedantam2015cider}. For GQA \citep{hudson2019gqa}, we report exact-match accuracy on the balanced test-dev split. To assess whether the learned hierarchy captures annotated part-whole structure, we also evaluate object and part localization and parent binding on PACO \citep{ramanathan2023paco}, with details provided in \cref{app:downstream,app:paco}.

\textbf{Control comparisons.}
To assess slot-count and OC grouping effects, we include a 35-slot flat DINOSAUR reference and an ungrouped frozen-backbone reference in the main COCO and GQA comparisons. Both use the same Perceiver-64 architecture and task fine-tuning recipe as the other models, with separately trained weights. The 35-slot reference matches the top-down all-level input count without introducing a hierarchy, and the backbone reference tests the usefulness of slot grouping. For more details, see \cref{app:downstream}.

\begin{wraptable}{R}{0.49\textwidth}
\centering
\captionsetup{font=small,justification=raggedright,singlelinecheck=false}
\caption{Downstream results with DINOv2-S. Values are mean $\pm$ sample standard deviation over three seeds. OC rows vary upstream seeds, and DINOv2-S varies task-training seeds. All rows use separately trained Perceiver-64 adapters and the same training recipe. Bold indicates the best OC result in each column.}
\label{tab:main}
\smallskip
\fontsize{8.5}{10}\selectfont
\setlength{\tabcolsep}{1.5pt}
\renewcommand{\arraystretch}{1.10}
\begin{tabular*}{\linewidth}{@{\extracolsep{\fill}}lccc@{}}
\toprule
Model & Input & \shortstack{COCO\\CIDEr $\uparrow$} & \shortstack{GQA\\Acc. $\uparrow$} \\
\midrule
DINOv2-S & Patches & $104.06{\pm}0.53$ & $49.28{\pm}0.25$ \\
\midrule
\multirow{2}{*}{DINOSAUR} & Objects (7) & $99.34{\pm}0.32$ & $47.45{\pm}0.21$ \\
 & Objects (35) & $99.21{\pm}0.55$ & $48.09{\pm}0.25$ \\
\multirow{3}{*}{\oursTD{}} & Objects (7) & $99.38{\pm}0.75$ & $47.65{\pm}0.23$ \\
 & Parts (28) & $\mathbf{100.16{\pm}0.32}$ & $\mathbf{48.63{\pm}0.07}$ \\
 & All (35) & $99.85{\pm}0.88$ & $48.41{\pm}0.39$ \\
\multirow{3}{*}{\oursBU{}} & Objects (7) & $98.74{\pm}0.74$ & $47.18{\pm}0.25$ \\
 & Parts (32) & $97.40{\pm}1.10$ & $47.52{\pm}0.24$ \\
 & All (39) & $97.19{\pm}1.45$ & $47.46{\pm}0.26$ \\
\bottomrule
\end{tabular*}
\end{wraptable}

\textbf{Results.}
\cref{tab:main_discovery,tab:main} summarize the main results with a DINOv2-S backbone. The hierarchical variants are close to seven-slot DINOSAUR in upstream segmentation: \oursBU{} achieves the highest FG-ARI, while DINOSAUR obtains the highest mBO. This suggests that the hierarchical models can form competitive masks, but they do not uniformly improve standard segmentation metrics. For downstream evaluation, \oursTD{} with all-level tokens reaches $99.85$ COCO CIDEr compared with $99.34$ for DINOSAUR. On DINOv2-S, top-down all-level slots reach 48.41\% GQA accuracy, compared with 47.45\% for seven flat slots and 48.09\% for the slot-count-matched flat model. In contrast, \oursBU{} with all-level tokens reaches $97.19$ CIDEr and $47.46\%$ GQA accuracy. Using all levels is not uniformly better than using objects alone, and parts-only inputs can be competitive with or stronger than all-level inputs. In particular, \oursTD{} parts alone obtain $100.16$ CIDEr and $48.63\%$ GQA accuracy. Additional backbones show the same direction of top-down all-level gains over seven flat slots on both tasks, with larger GQA gains, but bottom-up performance and the choice of which slot levels to use remain task-dependent (\cref{app:dinov3}).

\textbf{Slot-count and backbone controls.}
Increasing flat DINOSAUR from 7 to 35 slots leaves main-backbone captioning almost unchanged ($99.34$ versus $99.21$ CIDEr) and improves GQA from $47.45\%$ to $48.09\%$, narrowing the gap to \oursTD{} with all levels. Thus, some of the GQA advantage can be obtained by increasing the flat slot count and the capacity of the upstream OC model. However, the 35-slot control has lower FG-ARI and mBO than seven-slot DINOSAUR with all three backbones (\cref{tab:main_discovery,tab:additional_backbones}). The additional-backbone controls show larger, task-dependent slot-count effects (\cref{tab:additional_controls}). Ungrouped DINOv2-S features reach $104.06$ CIDEr and $49.28\%$ GQA accuracy. These scores exceed the best OC results, consistent with prior evidence that the benefits of OC methods depend on the evaluation setting \citep{kapl2026object}.

\begin{table}[t]
\centering
\captionsetup{font=small}
\caption{PACO part-whole diagnostic with DINOv2-S. Values are percentages, reported as mean $\pm$ sample standard deviation over three upstream seeds. Coverage uses IoU $0.5$. Binding accuracy is conditional on a localized part and a matched predicted parent. Binding coverage is the fraction of annotated parts meeting both conditions. Dashes indicate quantities that do not apply to the corresponding representation.}
\vspace{0.5em}
\label{tab:paco_main}
\small
\setlength{\tabcolsep}{5pt}
\renewcommand{\arraystretch}{1.10}
\begin{tabular}{@{}lcccc@{}}
\toprule
Model & Object mBO & Part coverage & Binding coverage & Binding accuracy \\
\midrule
DINOSAUR (7 slots) & $18.20{\pm}0.11$ & -- & -- & -- \\
DINOSAUR (35 slots) & $23.06{\pm}0.03$ & -- & -- & -- \\
\oursTD{} & $16.99{\pm}0.12$ & $2.21{\pm}0.13$ & $0.73{\pm}0.04$ & $96.19{\pm}1.99$ \\
\oursBU{} & $18.13{\pm}0.10$ & $1.74{\pm}0.10$ & $0.60{\pm}0.04$ & $97.68{\pm}1.37$ \\
\bottomrule
\end{tabular}
\end{table}

\textbf{Part-whole diagnostic.}
On PACO \citep{ramanathan2023paco}, \cref{tab:paco_main} shows that the hierarchical models provide explicit parent associations, but localization coverage remains limited. At IoU $0.5$, \oursTD{} and \oursBU{} localize $2.21\%$ and $1.74\%$ of annotated parts, respectively. Only $0.73\%$ and $0.60\%$ of annotated parts enter the binding calculation, within which parent-binding accuracy reaches $96.19\%$ and $97.68\%$, respectively. Object mBO does not improve over either flat baseline. Increasing DINOSAUR from 7 to 35 slots improves object mBO from $18.20$ to $23.06$, without adding explicit part-whole structure. Thus, explicit parent associations alone do not ensure broad coverage of annotated part-whole structure. Further results are provided in \cref{app:downstream,app:paco}.

\begin{figure}[t]
  \centering
  \includegraphics[width=0.98\linewidth]{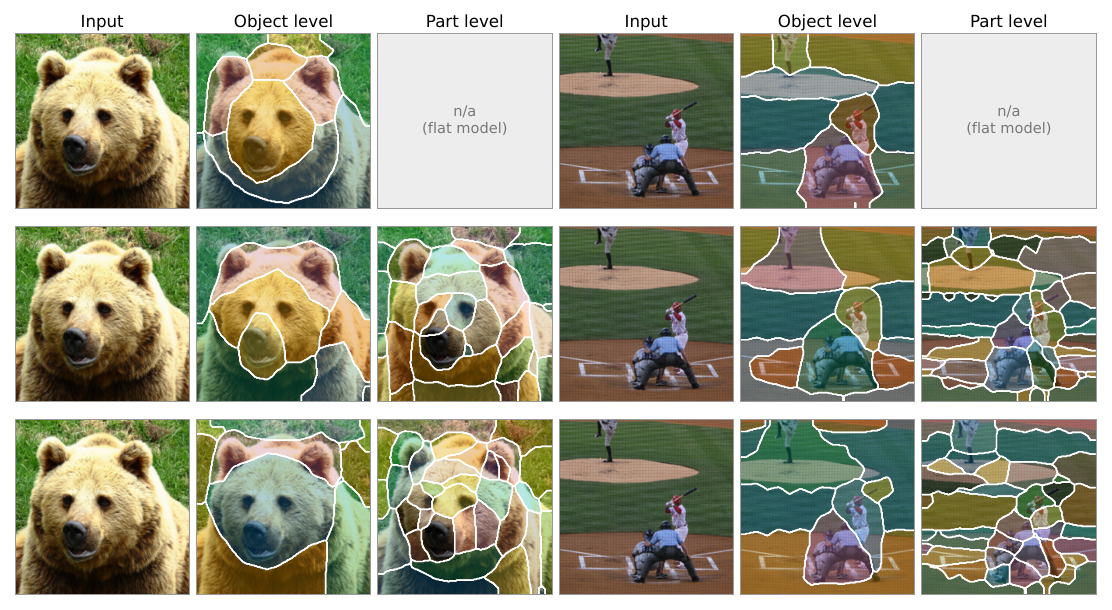}
  \caption{Examples of decoder-derived masks on two COCO scenes. Rows show, from top to bottom, DINOSAUR, \oursTD{}, and \oursBU{}, and from left to right, columns show the image, object-level masks, and part-level masks. Hierarchical models can produce visually plausible multi-level decompositions, but these decompositions do not necessarily correspond to a semantic parse tree.}
  \label{fig:qualitative}
\end{figure}

Qualitative masks, shown in \cref{fig:qualitative}, illustrate the masks corresponding to slots learned at each level. \oursTD{} conditions part discovery on object-level regions, while \oursBU{} groups part slots into objects. The decoded masks can form coherent regions, but their boundaries need not align across levels. These visualizations support the interpretation that the models learn multi-level groupings. At the same time, they also show why mask quality alone is not enough: a mask can correspond to a coherent visual region without encoding the semantic or relational information needed for captioning or question answering. Overall, the quantitative and qualitative results suggest that part-whole slots are useful as multi-level visual representations, but they do not establish that the learned hierarchy forms a semantic parse tree.

\suppressfloats[t]

\section{Discussion}
\label{sec:discussion}

Our results are consistent with recent hierarchical OC studies showing that the benefits of hierarchy can be strongly level- and objective-dependent \citep{madan2026hsa}. In particular, recent work reports trade-offs between coarse and fine hierarchy levels, suggesting that improving one level does not necessarily improve all others. In our reconstruction-based setting, part-level inputs can be useful without a consistent advantage from combining levels, while strong object-discovery performance does not necessarily translate into stronger downstream representations. This suggests that architectural hierarchy alone is insufficient to guarantee an advantage. Objectives that specify what information should be preserved, grouped, or abstracted across levels are a promising direction. Below, we discuss three factors that may underlie this gap.

\textbf{Segmentation ambiguity remains an issue.}
A visual scene admits many valid decompositions. A model may group pixels by object identity, semantic part, texture, color, depth, motion, or simply by what best reconstructs the chosen feature space \citep{locatello2020object}. A reconstruction loss alone does not specify which decomposition is the intended one, and even if the model can detect proper objects and parts, there is no control over what is recognized as ``object'' or ``part'', and at what granularity. This ambiguity already exists for object-level OC learning \citep{locatello2020object,seitzer2022bridging} and becomes more severe for part-whole hierarchies, where the desired level of detail is similarly task-dependent \citep{greff2020binding}. A chair can be represented as one object, as back, seat, and legs, or as many smaller surfaces and edges. Without an additional signal, none of these decompositions is uniquely preferred by the training objective.

\textbf{Object-like slots are not necessarily semantic objects.}
A slot can be useful for reconstruction or feature grouping without corresponding to a human-meaningful object or part \citep{seitzer2022bridging}, as Slot Attention is often described as soft K-means clustering \citep{locatello2020object}. This distinction matters for downstream tasks. Captioning and VQA require categories, attributes, relations, and functions. A mask over a coherent visual region does not guarantee that this information is represented in a useful form. This helps explain why \oursBU{} can achieve strong FG-ARI, even though its object-level slots do not improve on DINOSAUR in COCO captioning. Therefore, semantic grounding methods, including language-aligned OC models \citep{didolkar2025ctrl}, are complementary and preferred to purely reconstruction-based training.

\textbf{Static images may not provide enough signal.}
Humans and embodied agents do not learn object structure from isolated images alone. Motion, interaction, language, touch, proprioception, and task feedback all help determine what counts as a coherent object or part. Classic developmental studies emphasize the role of motion and spatial arrangement in object perception \citep{spelke1990principles}, and modern arguments for human-like learning similarly emphasize richer multimodal and interactive data \citep{lake2017building}. Static-image models can be useful, but a static-image signal without additional supervision is likely too weak to serve as the sole signal for learning a stable, semantic part-whole structure. Video-based models, weak supervision, language grounding, and online interactive settings are promising directions to alleviate this limitation \citep{elsayed2022savi++,wu2022slotformer,madan2026hsa}.

\textbf{Takeaway.}
Our work, together with recent work on hierarchical OC, shows that hierarchical structure can be incorporated into object-centric representations in several ways \citep{maus2025efficient,dai2026cgsa,madan2026hsa}. Our results suggest that the more difficult question is not how to create multiple slot levels, but how to ensure that those levels correspond to meaningful abstractions and that higher-level representations preserve the relevant information from their parts. In this view, hierarchy should specify not only which slots exist at each level, but also how information is bound, compressed, and communicated across levels. Reconstruction alone leaves much of this structure underdetermined. Together with recent evaluation studies of OC methods \citep{mamaghan2025exploring,kapl2026object,rubinstein2025donewithoc,baldassarre2024clustering}, our results suggest that current OC is better viewed as a useful inductive bias under constraints rather than as a complete vision model. Realizing its potential for part-whole reasoning will therefore require addressing these constraints through directions such as explicit cross-level objectives, semantic grounding, adaptive granularity, and richer temporal signals from videos or other task-dependent learning signals.

\section{Conclusion}
\label{sec:conclusion}
We study how explicit part-whole structure behaves in object-centric representations by comparing two complementary Slot Attention hierarchies. The resulting models provide useful multi-level representations, with modest captioning gains and clearer VQA gains. However, hierarchy does not by itself provide a general downstream advantage. The benefits depend on the architecture, representation level, task, and backbone. Evaluations on PACO further show that downstream usefulness can coexist with limited alignment to annotated part-whole structure. We therefore frame this work as a case study in the promise and limitations of current approaches to part-whole object-centric learning. Promising directions for part-whole visual reasoning include augmenting reconstruction-based learning with richer semantic and temporal signals, explicit cross-level objectives, adaptive granularity, and stronger relational structure.

\begin{ack}
This work was partially supported by the Wallenberg AI, Autonomous Systems and Software Program (WASP), funded by the Knut and Alice Wallenberg Foundation, and by the Helmholtz Foundation Model Initiative, supported by the Helmholtz Association.
The authors thank the Gauss Centre for Supercomputing e.V. (GCS) for providing computing time on the Supercomputer JUPITER at the Jülich Supercomputing Centre via the John von Neumann Institute for Computing. JUPITER is supported by the EuroHPC JU and GCS through funding by the European Commission, the German Federal Ministry of Research, Technology and Space, and the Ministry of Culture and Science of the State of North Rhine-Westphalia. The computations were also enabled by the Berzelius resource, provided by the Knut and Alice Wallenberg Foundation at the National Supercomputer Centre. Furthermore, the authors appreciate the computational resources provided by the National High Performance Computing Centre (www.nhr.kit.edu). The research presented is supported by the TUM Georg Nemetschek Institute Artificial
Intelligence for the Built World.
\end{ack}

{\small
\bibliography{references}
\bibliographystyle{unsrtnat}
}

\newpage
\appendix

\section{Architecture and Training Details}
\label{app:arch}

\subsection{Upstream Training}

\paragraph{Visual backbone.}
All OC encoders operate on frozen self-supervised ViT features. In our experiments, we use DINOv2 ViT-S/14 as the main frozen backbone. Images are resized with the aspect ratio preserved so that the shorter side is 224 pixels, then center-cropped to $224\times224$. This gives a $16\times16$ grid of patch features, i.e., $N=256$ patch tokens, and the feature dimension is $C=384$. We use ImageNet \citep{deng2009imagenet} normalization and random horizontal flipping during training. We also report additional results with DINOv3 ViT-S/16 and DINOv3 ViT-L/16 in \cref{app:dinov3}. At the same image resolution, these produce $14\times14=196$ patch tokens with feature dimensions $384$ and $1024$, respectively. The upstream training hyperparameters are summarized in \cref{tab:upstream_shared_hparams,tab:upstream_model_hparams}.

\paragraph{Training objective.}
Each Slot Attention level produces a set of slots and attention masks over its input tokens. Similar to DINOSAUR \citep{seitzer2022bridging}, we use a Patch-MLP decoder that reconstructs the original frozen backbone features for each hierarchy level. The upstream objective is the sum of mean-squared reconstruction losses across hierarchy levels.

\paragraph{Object discovery evaluation.}
We evaluate decoder-derived object masks on COCO instance annotations, excluding overlapping-instance pixels and images with no remaining foreground after cropping. FG-ARI and mBO are averaged over the remaining images.

\paragraph{Decoder.}
Following DINOSAUR \citep{seitzer2022bridging}, we use a 3-layer Patch-MLP decoder at each hierarchy level. The decoder maps each slot to patch-level feature predictions and alpha masks. The final reconstruction is obtained by alpha-compositing the slot-wise predictions over patches. All decoders reconstruct the original backbone feature space, including the second level of \oursBU{}, whose input tokens are part slots rather than image patches.

\paragraph{Slot relation encoder.}
After each Slot Attention module, we apply a shallow slot relation encoder. This is a single transformer encoder layer over the slots at the same hierarchy level. It allows slots to communicate and exchange information before being decoded or passed to the next hierarchy level \citep{hinton2023glom}. The slot relation encoder uses a standard transformer layer with four attention heads and an MLP with hidden dimension $4d$, where $d$ is the slot dimension.

\paragraph{DINOSAUR.}
DINOSAUR \citep{seitzer2022bridging} is used as the flat OC baseline. It uses the same frozen backbone features as the corresponding part-whole \ours{} models, but has only one Slot Attention level. We use $K=7$ slots for the baseline and $K=35$ for the slot-count control, with slot dimension $256$, hidden dimension $1024$, and three Slot Attention iterations in both cases. The full upstream model hyperparameters for DINOSAUR, \oursTD{}, and \oursBU{} are reported in \cref{tab:upstream_model_hparams}.

\paragraph{\oursTDb{}.}
\oursTD{} follows an object-to-parts hierarchy. The first Slot Attention module receives the full set of backbone patch features and produces $7$ object-level slots. The attention masks from this level are then used to mask the original backbone features, producing one masked feature set per object. A shared second Slot Attention module is applied to each masked object region and produces $4$ part slots per object. Therefore, the model produces $7$ object-level slots and $7\times4=28$ part-level slots. Both levels use three Slot Attention iterations and their own Patch-MLP decoder.

\paragraph{\oursBUb{}.}
\oursBU{} follows a parts-to-objects hierarchy. The first Slot Attention module receives the full set of backbone patch features and produces $32$ part-level slots. The second Slot Attention module takes these part slots as input and groups them into $7$ object-level slots. Both levels use three Slot Attention iterations and their own Patch-MLP decoder.

\subsection{Upstream datasets}

We train the upstream OC encoders on the image sets corresponding to the downstream evaluation datasets. For COCO \citep{lin2014microsoft}, we use the COCO 2017 training images and validate on the COCO 2017 validation images. For GQA \citep{hudson2019gqa}, we use a $90/10$ train/validation split of the GQA images. All images use the resize and center-crop preprocessing described above.

\begin{table}[t]
    \caption{Shared upstream training hyperparameters for DINOSAUR, \oursTD{}, and \oursBU{}.}
    \vspace{5pt}
    \label{tab:upstream_shared_hparams}
    \centering
    \begin{tabular*}{\linewidth}{@{\extracolsep{\fill}}ll@{}}
        \toprule
        Hyperparameter & Value \\
        \midrule
        Optimizer & AdamW \\
        Learning rate & $3\times10^{-4}$ \\
        Schedule & cosine \\
        Warmup steps & 10{,}000 \\
        Training steps & 300{,}000 \\
        Batch size & 128 \\
        Gradient clipping & 0.1 \\
        \bottomrule
    \end{tabular*}
\end{table}

\begin{table}[t]
    \caption{Upstream model hyperparameters.$^\dagger$}
    \vspace{5pt}
    \label{tab:upstream_model_hparams}
    \centering
    \begin{tabular*}{\linewidth}{@{\extracolsep{\fill}}lccc@{}}
        \toprule
        Hyperparameter & DINOSAUR & \oursTD{} & \oursBU{} \\
        \midrule
        Number of levels & 1 & 2 & 2 \\
        Slots per level & 7 or 35 & 7, 4 per object & 32, 7 \\
        Total slots & 7 or 35 & 35 & 39 \\
        Slot dimensions & 256 & 256, 128 & 128, 256 \\
        Slot hidden dimensions & 1024 & 1024, 512 & 512, 1024 \\
        Slot iterations & 3 & 3, 3 & 3, 3 \\
        Slot relation encoder & yes & yes, yes & yes, yes \\
        Slot relation heads & 4 & 4, 4 & 4, 4 \\
        Slot relation layers & 1 & 1, 1 & 1, 1 \\
        Decoder hidden dims & 2048 & 2048, 1024 & 1024, 2048 \\
        \bottomrule
        \multicolumn{4}{l}{\footnotesize{$^\dagger$ For part-whole \ours{} models, comma-separated values correspond to the hierarchy levels.}} \\
        \bottomrule
    \end{tabular*}
\end{table}

\begin{table}[t]
    \caption{Shared downstream training hyperparameters.}
    \vspace{5pt}
    \label{tab:downstream_shared_hparams}
    \centering
    \begin{tabular*}{\linewidth}{@{\extracolsep{\fill}}ll@{}}
        \toprule
        Hyperparameter & Value \\
        \midrule
        Language model & Phi-3-mini-4k-instruct \\
        Visual-token adapter & Perceiver-64 \\
        Optimizer & AdamW \\
        Learning rate & $1\times10^{-5}$ \\
        Weight decay & $1\times10^{-6}$ \\
        Schedule & cosine with $5\%$ linear warmup \\
        Training epochs & 4 \\
        Batch size & 16 \\
        \bottomrule
    \end{tabular*}
\end{table}

\section{Downstream Training Details}
\label{app:downstream}

\paragraph{VLM setup.}
For downstream evaluation, we use the BLIP-3 \citep{xue2024xgen} setup with Phi-3-mini-4k-instruct \citep{abdin2024phi3} as the language model. The original visual encoder is replaced by the frozen OC encoder. We initialize the language weights from the converted BLIP-3 instruction-tuned checkpoint and initialize the visual adapter separately. During downstream training, the OC encoder remains frozen, while the language model and the visual-token adapter are fine-tuned. The downstream model is trained with a supervised fine-tuning objective, using cross-entropy loss only on the target answer tokens.

\paragraph{Visual tokens.}
For DINOSAUR, the visual tokens are the $7$ baseline slots or the $35$ control slots. For \oursTD{}, the object-level setting uses the $7$ first-level object slots, while the all-level setting concatenates the $7$ object slots and the $28$ part slots, giving $35$ visual tokens. For \oursBU{}, the object-level setting uses the $7$ second-level object slots, while the all-level setting concatenates the $32$ part slots and the $7$ object slots, giving $39$ visual tokens. The parts-only references use the $28$ part slots of \oursTD{} or the $32$ part slots of \oursBU{}.

\paragraph{Visual-token adapter.}
We use the same Perceiver adapter architecture \citep{jaegle2021perceiver} with 64 learned queries for every model and token choice, with weights trained separately for each run. The queries cross-attend to the selected slots and produce a fixed-length sequence of visual tokens. The adapter has six layers, $16$ attention heads, head dimension $96$, feed-forward multiplier $4$, and a final linear projection from the visual feature width to the Phi-3 hidden dimension $3072$. The visual width is $256$ for slot inputs and $384$ or $1024$ for the ungrouped backbone references. We project the 128-dimensional part slots to the 256-dimensional object-slot width for parts-only and all-level inputs. This projection is trained jointly with the adapter and language model while the OC encoder remains frozen.

\paragraph{Training data.}
For COCO caption fine-tuning, we pair train2017 images \citep{lin2014microsoft} with their official human-written captions \citep{chen2015cococaptions}. We reserve 5\% of training images for validation and use all 5{,}000 val2017 images for evaluation. The prompt is ``Describe this image in one sentence.'' Following LLaVA-1.5's visual instruction-tuning format \citep{liu2024improved}, we use its GQA subset \citep{hudson2019gqa} for question answering, with each question as the instruction and its answer string as the target. We also reserve 5\% of these images for validation. The supervised training, validation, and evaluation images are disjoint for each task. Examples use the Phi-3 chat template, with loss applied only to assistant answers and their end token.

\paragraph{Optimization.}
We fine-tune the language model together with the visual-token adapter while keeping the OC encoder frozen. We adopt the supervised fine-tuning formulation of BLIP-3 \citep{xue2024xgen} and use AdamW, a cosine learning-rate schedule, and a $5\%$ linear warmup. We use the same optimization settings across models and token choices, as summarized in \cref{tab:downstream_shared_hparams}. We evaluate the final checkpoint after four epochs, using greedy decoding with a maximum of 64 new tokens.

\paragraph{Control and part-whole diagnostic details.}
The 35-slot flat control uses the corresponding seven-slot baseline's pretraining recipe with only the slot count changed. Its slot count matches the top-down all-level input but not the bottom-up input of 39 slots. The ungrouped reference uses all frozen backbone patch tokens, whose widths differ from the slot inputs. Equal Perceiver output length therefore does not equalize input information or model capacity. On the 2{,}183-image intersection of Parts and Attributes of Common Objects (PACO) test annotations and COCO val2017 \citep{ramanathan2023paco}, we also evaluate object and part localization and annotated-parent binding. For mBO, each annotated mask independently takes its highest IoU over predicted masks at the corresponding level. We average these maxima within each image, then across eligible images. For localization coverage and parent binding, we use class-agnostic one-to-one Hungarian matching to maximize total IoU, then apply IoU thresholds of 0.25 and 0.5 to the matched pairs. We use decoder-alpha argmax masks and exclude crowd, ignored, and empty annotations after cropping, as well as parts without an annotated parent. For \oursTD{}, each part inherits the parent index defined by the architecture. For \oursBU{}, its parent is the object slot with the largest second-level attention assignment, while object and part masks are decoded separately. Binding accuracy measures correct parent identity among localized parts whose predicted parent is also matched, and is reported with binding coverage. Both the seven-slot and 35-slot flat baselines participate in object localization but have no explicit part level or parent map. This diagnostic measures alignment of decoder-derived masks from coarse feature grids with cropped annotations.

\section{Additional Backbone Results}
\label{app:dinov3}

We additionally evaluate the models with DINOv3 ViT-S/16 and DINOv3 ViT-L/16 \citep{simeoni2025dinov3} backbones in \cref{tab:additional_backbones}. These experiments test whether the main observations are sensitive to the frozen visual backbone and its size. With both backbones, \oursTD{} with all-level inputs obtains the strongest OC result on both tasks. Its captioning gains over DINOSAUR remain modest ($97.17$ versus $96.16$ CIDEr for DINOv3-S and $109.02$ versus $108.67$ for DINOv3-L), while GQA improves by $3.00$ and $2.32$ percentage points, respectively. Parts-only top-down inputs are also competitive, and bottom-up results depend on the task and backbone. Overall, these results support useful top-down gains without a general advantage from every hierarchical configuration.

\begin{table}[t]
    \caption{Additional backbone results. Segmentation uses COCO instance annotations. Parentheses in panel (a) indicate flat slot counts. Every configuration reports mean $\pm$ standard deviation over three upstream seeds. Bold indicates the best result for each backbone and metric.}
    \vspace{0.5em}
    \label{tab:additional_backbones}
    \centering
\small
\setlength{\tabcolsep}{2pt}
\renewcommand{\arraystretch}{1.10}
\textbf{(a) Unsupervised Object Discovery}
\par\smallskip
\begin{tabular*}{\linewidth}{@{\extracolsep{\fill}}lcc@{}}
\toprule
Model & FG-ARI $\uparrow$ & mBO $\uparrow$ \\
\midrule
\multicolumn{3}{@{}l}{\textit{DINOv3-S}} \\
\addlinespace[2pt]
DINOSAUR (7) & $36.71{\pm}0.22$ & $\mathbf{28.93{\pm}0.02}$ \\
DINOSAUR (35) & $19.92{\pm}0.08$ & $24.34{\pm}0.12$ \\
\oursTD{} & $36.15{\pm}0.28$ & $27.83{\pm}0.06$ \\
\oursBU{} & $\mathbf{37.42{\pm}0.32}$ & $28.34{\pm}0.19$ \\
\midrule
\multicolumn{3}{@{}l}{\textit{DINOv3-L}} \\
\addlinespace[2pt]
DINOSAUR (7) & $39.77{\pm}0.16$ & $\mathbf{29.65{\pm}0.05}$ \\
DINOSAUR (35) & $20.74{\pm}0.20$ & $24.33{\pm}0.13$ \\
\oursTD{} & $37.63{\pm}0.19$ & $27.73{\pm}0.13$ \\
\oursBU{} & $\mathbf{40.32{\pm}0.72}$ & $28.51{\pm}0.14$ \\
\bottomrule
\end{tabular*}
\par\medskip
\textbf{(b) Downstream tasks}
\par\smallskip
\begin{tabular*}{\linewidth}{@{\extracolsep{\fill}}lccc@{}}
\toprule
Model & Input & COCO CIDEr $\uparrow$ & GQA Acc. $\uparrow$ \\
\midrule
\multicolumn{4}{@{}l}{\textit{DINOv3-S}} \\
\addlinespace[2pt]
DINOSAUR & Objects (7) & $96.16{\pm}0.41$ & $45.48{\pm}0.43$ \\
\midrule
\multirow{3}{*}{\oursTD{}} & Objects (7) & $95.85{\pm}0.56$ & $45.92{\pm}0.37$ \\
 & Parts (28) & $96.48{\pm}0.36$ & $48.27{\pm}0.27$ \\
 & All (35) & $\mathbf{97.17{\pm}0.36}$ & $\mathbf{48.48{\pm}0.13}$ \\
\midrule
\multirow{3}{*}{\oursBU{}} & Objects (7) & $95.32{\pm}0.63$ & $46.18{\pm}0.50$ \\
 & Parts (32) & $96.13{\pm}0.55$ & $47.55{\pm}0.52$ \\
 & All (39) & $95.89{\pm}0.62$ & $47.46{\pm}0.73$ \\
\midrule
\multicolumn{4}{@{}l}{\textit{DINOv3-L}} \\
\addlinespace[2pt]
DINOSAUR & Objects (7) & $108.67{\pm}0.60$ & $47.11{\pm}0.48$ \\
\midrule
\multirow{3}{*}{\oursTD{}} & Objects (7) & $106.59{\pm}0.69$ & $47.31{\pm}0.15$ \\
 & Parts (28) & $108.74{\pm}1.03$ & $49.35{\pm}0.18$ \\
 & All (35) & $\mathbf{109.02{\pm}1.41}$ & $\mathbf{49.43{\pm}0.24}$ \\
\midrule
\multirow{3}{*}{\oursBU{}} & Objects (7) & $107.39{\pm}0.98$ & $48.12{\pm}0.46$ \\
 & Parts (32) & $106.87{\pm}0.57$ & $48.33{\pm}0.21$ \\
 & All (39) & $106.53{\pm}0.61$ & $48.13{\pm}0.26$ \\
\bottomrule
\end{tabular*}
\end{table}

\clearpage
\subsection{Additional controls}
\cref{tab:additional_controls} compares all configurations on one seed. Increasing the flat slot count improves GQA from $45.86\%$ to $47.72\%$ with DINOv3-S and from $46.97\%$ to $49.12\%$ with DINOv3-L, narrowing the top-down all-level advantage. Its captioning effect differs across backbones, decreasing CIDEr from $96.18$ to $93.42$ with DINOv3-S and increasing it from $108.46$ to $108.78$ with DINOv3-L. Thus, the small main-backbone captioning effect does not generalize to every backbone. Ungrouped features remain stronger in these comparisons, subject to the input-information and capacity differences described in \cref{app:downstream}.

\begin{table}[!htbp]
    \caption{Slot-count and ungrouped-feature comparisons on both tasks with additional backbones. The results are reported on one seed. Bold indicates the best OC result for each backbone and metric.}
    \label{tab:additional_controls}
    \centering
\small
\setlength{\tabcolsep}{2pt}
\renewcommand{\arraystretch}{1.10}
\begin{tabular*}{\linewidth}{@{\extracolsep{\fill}}lccc@{}}
\toprule
Model & Visual input & COCO CIDEr $\uparrow$ & GQA Acc. $\uparrow$ \\
\midrule
\multicolumn{4}{@{}l}{\textit{DINOv3-S}} \\
\addlinespace[2pt]
DINOv3-S & Patches & $100.36$ & $49.22$ \\
\midrule
\multirow{2}{*}{DINOSAUR} & Objects (7) & $96.18$ & $45.86$ \\
 & Objects (35) & $93.42$ & $47.72$ \\
\midrule
\multirow{3}{*}{\oursTD{}} & Objects (7) & $95.94$ & $45.56$ \\
 & Parts (28) & $96.71$ & $48.31$ \\
 & All (35) & $\mathbf{97.49}$ & $\mathbf{48.43}$ \\
\midrule
\multirow{3}{*}{\oursBU{}} & Objects (7) & $94.69$ & $45.61$ \\
 & Parts (32) & $96.65$ & $48.14$ \\
 & All (39) & $96.58$ & $48.28$ \\
\midrule
\multicolumn{4}{@{}l}{\textit{DINOv3-L}} \\
\addlinespace[2pt]
DINOv3-L & Patches & $117.06$ & $50.97$ \\
\midrule
\multirow{2}{*}{DINOSAUR} & Objects (7) & $108.46$ & $46.97$ \\
 & Objects (35) & $108.78$ & $49.12$ \\
\midrule
\multirow{3}{*}{\oursTD{}} & Objects (7) & $106.79$ & $47.14$ \\
 & Parts (28) & $109.86$ & $49.56$ \\
 & All (35) & $\mathbf{110.53}$ & $\mathbf{49.66}$ \\
\midrule
\multirow{3}{*}{\oursBU{}} & Objects (7) & $106.49$ & $48.65$ \\
 & Parts (32) & $106.57$ & $48.10$ \\
 & All (39) & $106.16$ & $48.43$ \\
\bottomrule
\end{tabular*}
\end{table}

\clearpage
\begingroup
\setlength{\floatsep}{8pt plus 2pt minus 2pt}
\setlength{\textfloatsep}{8pt plus 2pt minus 2pt}
\setlength{\intextsep}{8pt plus 2pt minus 2pt}
\section{Part-Whole Localization and Binding}
\label{app:paco}
\cref{tab:paco_overlap,tab:paco_025,tab:paco_050} report the PACO diagnostic defined in \cref{app:downstream}. Hierarchical models do not improve object mBO over either flat baseline. The 35-slot control has the highest object mBO with all three backbones. At IoU $0.5$, the two hierarchical models' parent-binding accuracy exceeds $95\%$ on average for every backbone, but binding coverage remains below $1\%$. At IoU $0.25$, coverage increases while binding accuracy decreases. The explicit parent associations are therefore often correct within the small localized subset, but this does not establish broad part-whole localization or semantic part recognition.

\begin{table}[!htbp]
    \captionsetup{font=footnotesize,skip=3pt}
    \caption{PACO object and part mean best overlap. All values are percentages, reported as mean $\pm$ sample standard deviation over three upstream seeds. Dashes indicate quantities that do not apply to the corresponding representation.}
    \label{tab:paco_overlap}
    \centering
    \footnotesize
    \setlength{\tabcolsep}{5pt}
    \renewcommand{\arraystretch}{0.95}
\begin{tabular*}{\linewidth}{@{\extracolsep{\fill}}lccc@{}}
\toprule
Backbone & Model & Object mBO & Part mBO \\
\midrule
DINOv2-S & DINOSAUR (7 slots) & $18.20{\pm}0.11$ & -- \\
DINOv2-S & DINOSAUR (35 slots) & $23.06{\pm}0.03$ & -- \\
DINOv2-S & \oursTD{} & $16.99{\pm}0.12$ & $12.66{\pm}0.03$ \\
DINOv2-S & \oursBU{} & $18.13{\pm}0.10$ & $12.36{\pm}0.11$ \\
\midrule
DINOv3-S & DINOSAUR (7 slots) & $18.51{\pm}0.20$ & -- \\
DINOv3-S & DINOSAUR (35 slots) & $23.88{\pm}0.17$ & -- \\
DINOv3-S & \oursTD{} & $17.56{\pm}0.05$ & $12.23{\pm}0.04$ \\
DINOv3-S & \oursBU{} & $18.06{\pm}0.23$ & $12.09{\pm}0.03$ \\
\midrule
DINOv3-L & DINOSAUR (7 slots) & $18.86{\pm}0.13$ & -- \\
DINOv3-L & DINOSAUR (35 slots) & $23.89{\pm}0.19$ & -- \\
DINOv3-L & \oursTD{} & $17.51{\pm}0.09$ & $12.43{\pm}0.05$ \\
DINOv3-L & \oursBU{} & $18.06{\pm}0.15$ & $12.24{\pm}0.19$ \\
\bottomrule
\end{tabular*}
\end{table}

\begin{table}[!htbp]
    \captionsetup{font=footnotesize,skip=3pt}
    \caption{PACO localization and parent binding at IoU 0.25. All values are percentages, reported as mean $\pm$ sample standard deviation over three upstream seeds. Dashes indicate quantities that do not apply to the corresponding representation.}
    \label{tab:paco_025}
    \centering
    \footnotesize
    \setlength{\tabcolsep}{5pt}
    \renewcommand{\arraystretch}{0.95}
\begin{tabular*}{\linewidth}{@{\extracolsep{\fill}}lccccc@{}}
\toprule
Backbone & Model & Object cov. & Part cov. & Binding cov. & Binding acc. \\
\midrule
DINOv2-S & DINOSAUR (7 slots) & $13.30{\pm}0.21$ & -- & -- & -- \\
DINOv2-S & DINOSAUR (35 slots) & $26.24{\pm}0.20$ & -- & -- & -- \\
DINOv2-S & \oursTD{} & $12.48{\pm}0.06$ & $10.44{\pm}0.05$ & $5.61{\pm}0.08$ & $91.55{\pm}0.90$ \\
DINOv2-S & \oursBU{} & $13.34{\pm}0.17$ & $10.25{\pm}0.10$ & $5.68{\pm}0.25$ & $91.93{\pm}1.04$ \\
\midrule
DINOv3-S & DINOSAUR (7 slots) & $13.25{\pm}0.13$ & -- & -- & -- \\
DINOv3-S & DINOSAUR (35 slots) & $26.79{\pm}0.27$ & -- & -- & -- \\
DINOv3-S & \oursTD{} & $12.39{\pm}0.07$ & $9.91{\pm}0.04$ & $5.39{\pm}0.09$ & $90.41{\pm}0.64$ \\
DINOv3-S & \oursBU{} & $12.95{\pm}0.15$ & $9.75{\pm}0.15$ & $5.25{\pm}0.08$ & $91.78{\pm}0.27$ \\
\midrule
DINOv3-L & DINOSAUR (7 slots) & $13.60{\pm}0.06$ & -- & -- & -- \\
DINOv3-L & DINOSAUR (35 slots) & $26.57{\pm}0.12$ & -- & -- & -- \\
DINOv3-L & \oursTD{} & $12.55{\pm}0.24$ & $10.13{\pm}0.12$ & $5.66{\pm}0.11$ & $91.59{\pm}1.28$ \\
DINOv3-L & \oursBU{} & $13.33{\pm}0.04$ & $10.06{\pm}0.32$ & $5.65{\pm}0.11$ & $92.10{\pm}0.24$ \\
\bottomrule
\end{tabular*}
\end{table}

\begin{table}[!htbp]
    \captionsetup{font=footnotesize,skip=3pt}
    \caption{PACO localization and parent binding at IoU 0.5. All values are percentages, reported as mean $\pm$ sample standard deviation over three upstream seeds. Dashes indicate quantities that do not apply to the corresponding representation.}
    \label{tab:paco_050}
    \centering
    \footnotesize
    \setlength{\tabcolsep}{5pt}
    \renewcommand{\arraystretch}{0.95}
\begin{tabular*}{\linewidth}{@{\extracolsep{\fill}}lccccc@{}}
\toprule
Backbone & Model & Object cov. & Part cov. & Binding cov. & Binding acc. \\
\midrule
DINOv2-S & DINOSAUR (7 slots) & $5.82{\pm}0.19$ & -- & -- & -- \\
DINOv2-S & DINOSAUR (35 slots) & $7.03{\pm}0.16$ & -- & -- & -- \\
DINOv2-S & \oursTD{} & $4.69{\pm}0.10$ & $2.21{\pm}0.13$ & $0.73{\pm}0.04$ & $96.19{\pm}1.99$ \\
DINOv2-S & \oursBU{} & $5.82{\pm}0.22$ & $1.74{\pm}0.10$ & $0.60{\pm}0.04$ & $97.68{\pm}1.37$ \\
\midrule
DINOv3-S & DINOSAUR (7 slots) & $6.22{\pm}0.21$ & -- & -- & -- \\
DINOv3-S & DINOSAUR (35 slots) & $7.69{\pm}0.29$ & -- & -- & -- \\
DINOv3-S & \oursTD{} & $5.35{\pm}0.03$ & $2.15{\pm}0.06$ & $0.75{\pm}0.03$ & $95.36{\pm}3.46$ \\
DINOv3-S & \oursBU{} & $5.90{\pm}0.12$ & $1.69{\pm}0.10$ & $0.61{\pm}0.05$ & $97.04{\pm}0.78$ \\
\midrule
DINOv3-L & DINOSAUR (7 slots) & $6.43{\pm}0.23$ & -- & -- & -- \\
DINOv3-L & DINOSAUR (35 slots) & $8.10{\pm}0.18$ & -- & -- & -- \\
DINOv3-L & \oursTD{} & $5.17{\pm}0.12$ & $2.27{\pm}0.14$ & $0.84{\pm}0.14$ & $95.67{\pm}1.11$ \\
DINOv3-L & \oursBU{} & $5.90{\pm}0.07$ & $1.70{\pm}0.13$ & $0.66{\pm}0.06$ & $98.79{\pm}1.42$ \\
\bottomrule
\end{tabular*}
\end{table}

\clearpage
\endgroup
\section{Additional Qualitative Results}
\label{app:qual}

\begin{figure}[!htbp]
  \centering
  \includegraphics[width=0.96\linewidth]{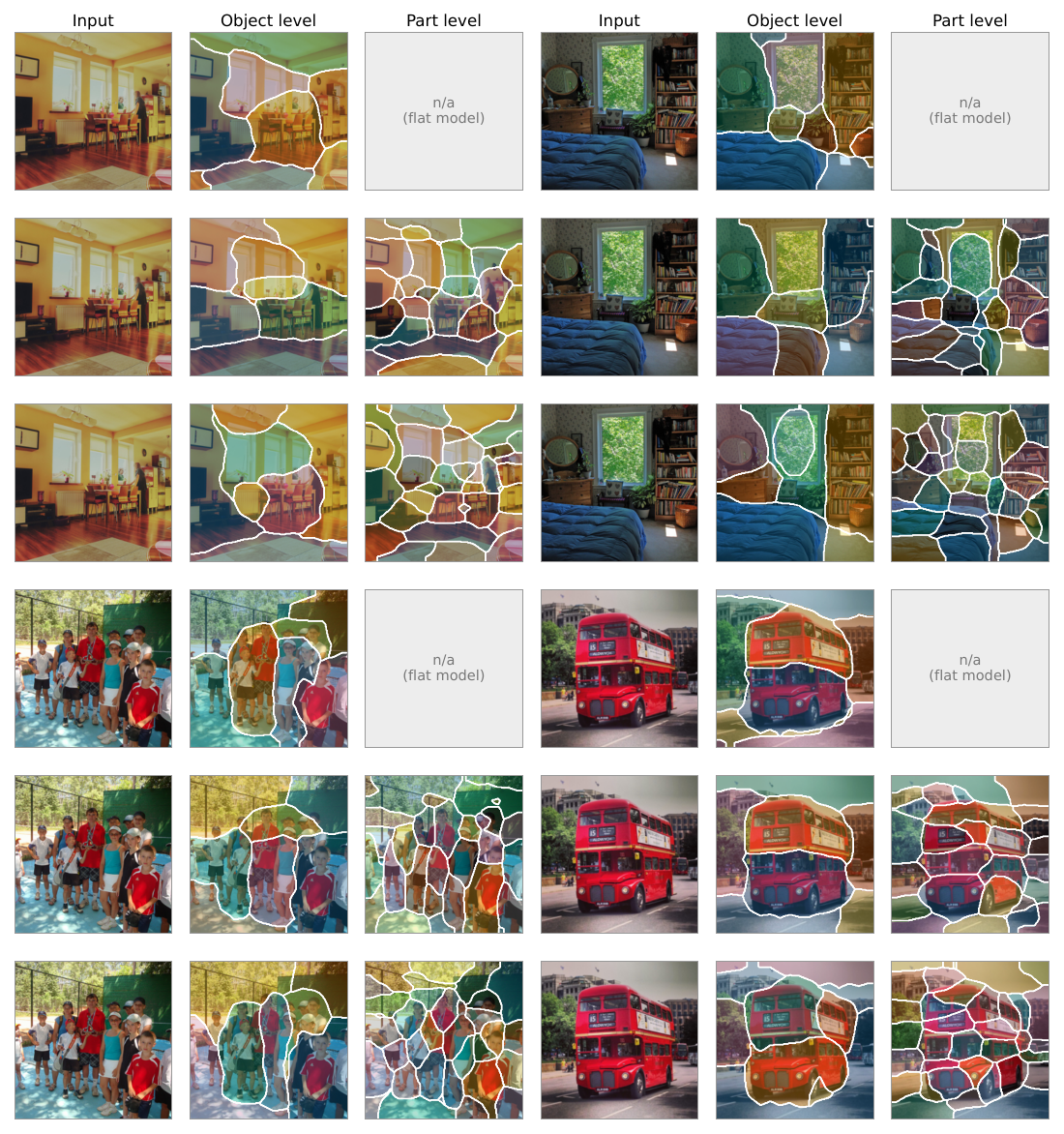}
  \caption{Additional decoder-derived masks for COCO scenes. For each image, rows show, from top to bottom, DINOSAUR, \oursTD{}, and \oursBU{}.}
  \label{fig:qual_additional}
\end{figure}

Additional qualitative masks are provided in \cref{fig:qual_additional}.

\end{document}